\documentclass[utf8]{frontiersSCNS} 

\usepackage{url,hyperref,lineno,microtype}
\usepackage[onehalfspacing]{setspace}

\usepackage[utf8]{inputenc} 
\usepackage[T1]{fontenc}    
\usepackage{booktabs}       
\usepackage{amssymb,amsmath,amsthm, amsfonts}
\usepackage{tabularx}
\usepackage{graphicx}
\usepackage{multirow}
\usepackage{titlesec}
\usepackage{caption} 
\usepackage[ruled,noline]{algorithm2e}
\usepackage{algorithmic}
\usepackage{mathtools}

\titlespacing*{\section}
{0pt}{8pt}{5pt}
\titlespacing*{\subsection}
{0pt}{8pt}{5pt}
\titlespacing*{\subsubsection}
{0pt}{7pt}{4pt}

\def\firstAuthorLast{Gao {et~al.}} 
\def\Authors{Yuyang Gao\,$^{1}$, Giorgio A. Ascoli\,$^{2}$ and Liang Zhao\,$^{1,*}$}
\def\Address{$^{1}$
Department of Information Sciences and Technology,
George Mason University,
Fairfax, VA, United States\\
$^{2}$
Center for Neural Informatics, Bioengineering Department, and Krasnow Institute for Advanced Study,
George Mason University,
Fairfax, VA, United States
}
\def\corrAuthor{Liang Zhao}
\def\corrEmail{lzhao9@gmu.edu}

\newtheorem{remark}{Remark}

\theoremstyle{definition}
\newtheorem{definition}{Definition}

\begin{document}
\onecolumn
\firstpage{1}

\title[Interpretable and Efficient Learning with BEAN Regularization]{BEAN: Interpretable and Efficient Learning with Biologically-Enhanced Artificial Neuronal Assembly Regularization} 

\author[\firstAuthorLast ]{\Authors} 
\address{} 
\correspondance{} 

\extraAuth{}

\maketitle

\begin{abstract}
Deep neural networks (DNNs) are known for extracting useful information from large amounts of data.
However, the representations learned in DNNs are typically hard to interpret, especially in dense layers.
One crucial issue of the classical DNN model such as multilayer perceptron (MLP) is that neurons in the same layer of DNNs are conditionally independent of each other, which makes co-training and emergence of higher modularity difficult.
In contrast to DNNs, biological neurons in mammalian brains display substantial dependency patterns. 
Specifically, biological neural networks encode representations by so-called neuronal assemblies: groups of neurons interconnected by strong synaptic interactions and sharing joint semantic content. The resulting population coding is essential for human cognitive and mnemonic processes.
Here, we propose a novel Biologically Enhanced Artificial Neuronal assembly (BEAN) regularization \footnote{Please find the source code at \url{https://drive.google.com/file/d/115bNbyiXj-Ip1RMExj3c18a6hliiNneo/view?usp=sharing.}} to model neuronal correlations and dependencies, inspired by cell assembly theory from neuroscience.
Experimental results show that BEAN enables the formation of interpretable neuronal functional clusters and consequently promotes a sparse, memory/computation-efficient network without loss of model performance.
Moreover, our few-shot learning experiments demonstrate that BEAN could also enhance the generalizability of the model when training samples are extremely limited.
\end{abstract}

\section{Introduction}

Deep neural networks (DNNs) are known for extracting useful information from a large amount of data [\cite{bengio2013representation}].
Despite the success and popularity of DNNs in a wide variety of fields, including computer vision [\cite{krizhevsky2012imagenet, he2016deep}] and natural language processing [\cite{collobert2008unified,young2018recent}],
there are still many drawbacks and limitations of modern DNNs, including lack of interpretability [\cite{zhang2018visual}], the requirement of large data [\cite{kimura2018few}], and post selection on complex model architecture [\cite{zheng2016mobile, zheng2016challenges}].
Specifically, the representations learned in DNNs are typically hard to interpret, especially in dense (fully connected) layers.
Despite recent attempts to build intrinsically more  interpretable convolutional units [\cite{zhang2018visual, sabour2017dynamic}], the  exploration of learned representations in the dense layer has remained limited. 
In fact, dense layers are the fundamental and critical component of most  state-of-the-art DNNs, which are typically used for the late stage of the network's computation, akin to the inference and decision-making processes  [\cite{krizhevsky2012imagenet, simonyan2014very, he2016deep}]. Thus improving  the interpretability of the dense layer representation is crucial if we are to fully understand and exploit the power of DNNs.

However, interpreting the representations learned in dense layers of DNNs is typically a very challenging task. 
One crucial issue of the classical DNN model such as multilayer perceptron (MLP) is that neurons in the same layer of DNNs are conditionally independent of each other, as dense layers in MLP are typically activated by  all-to-all feed-forward neuron activity and trained by all-to-all feedback weight adjustment. 
In this comprehensively `vertical' connectivity, every node is independent and abstracted `out of the context' of the other nodes. 
This issue limits the analysis of the representation learned in DNNs to single-unit level, as opposed to the higher modularity in principle afforded by neuron population coding.
Moreover, recent studies on single unit importance seem to suggest that individually selective units may have little correlation with overall network performance [\cite{morcos2018importance, zhou2018revisiting}].
Specifically, [\cite{morcos2018importance, zhou2018revisiting}] conducted unit-level ablation experiments on CNNs trained on large scale image datasets and found that ablating any individual unit does not hurt overall classification accuracy. 

On the other hand, 
understanding the complex patterns of neuron correlations in biological neural networks (BNNs) has long been a subject of intense interest for neuroscience researchers.
Circuitry blueprints in the real brain are `filtered' by the physical requirements of axonal projections and the consequent need to minimize cable while maximizing connections.
One could naively expect that the non-all-to-all limitations imposed in natural neural systems would be detrimental to their computational power. Instead, it makes them superiorly efficient and allows cell assemblies to emerge.
Neuronal assemblies or cell assemblies [\cite{hebb1949organization}] can be described as groups of neurons interconnected by strong synaptic interactions and sharing joint semantic content. The resulting population coding is essential for human cognitive and mnemonic processes [\cite{braitenberg1978cell}]. 


In this paper, we bridge such a crucial gap between DNNs and BNNs by modeling the neuron correlations within each layer of DNNs.
Leveraging biologically inspired learning rules in neuroscience and graph theory, we propose a novel Biologically-Enhanced Artificial Neuronal assembly (BEAN) regularization that can enforce dependencies among neurons in dense layers of DNNs without substantially altering the conventional architecture. The resultant advantages are threefold:

\begin{itemize}

\item \textbf{Enhancing interpretability and modularity at the neuron population level.}
Modeling neural correlations and dependencies  allows us to better interpret and visualize the learned representation in hidden layers at the neuron population level instead of the single neuron level.
Both qualitative and quantitative analyses show that BEAN enables the formations of identifiable neuronal assembly patterns in the hidden layers,  enhancing the modularity and interpretability of the DNN representations. 
\item \textbf{Promoting jointly sparse and efficient encoding of rich semantic correlation among neurons.}
Here, we show that BEAN can promote jointly sparse and efficient encoding of rich semantic correlation among neurons in DNNs similar to connection patterns in BNNs.
BEAN enables the model to parsimoniously leverage available neurons and possible connections through modeling structural correlation, yielding both connection-level and neuron-level sparsity in the dense layers. 
Experimental results show that BEAN not only enables the formation of neuronal functional clusters that encode rich semantic correlation, but also allows the model to achieve state-of-the-art memory/computational efficiency without loss of model performance.

\item \textbf{Improving model generalizability with few training samples.}
Humans and animals can learn and generalize to new concepts with just a few trials of learning, while DNNs generally perform poorly on such tasks.
Current few-shot learning techniques in deep learning still rely heavily on a large amount of additional knowledge to work well. 
For example, transfer-learning-based methods typically leverage a model pre-trained with a large amount of data [\cite{xian2018zero,socher2013zero}], and meta-learning-based methods require a large number of additional side tasks [\cite{finn2017model,snell2017prototypical}].
Here we explore BEAN with a substantially more challenging \textit{few-shot learning from scratch} task first studied by [\cite{kimura2018few}], where no additional knowledge is provided aside from a few training observations. Extensive experiments show that BEAN has a significant advantage in improving model generalizability over conventional techniques.

\end{itemize}


\section{Biologically-Enhanced Artificial Neuronal Assembly Regularization}

This section describes the overall objective of Biologically-Enhanced Artificial Neuronal Assembly (BEAN) regularization as well as the implementation of BEAN on DNNs, as \textit{Layer-wise Neuron Correlation and Co-activation Divergence} to model the implicit  dependencies between neurons within the same layer.

\subsection{Layer-wise Neuron Co-activation Divergence}
Due to the physical restrictions imposed by dendrites and axons [\cite{rivera2014wiring}] and for energy efficiency, biological neural systems are ``parsimonious'' and can only afford to form a limited number of connections between neurons. The neuron connectivity patterns of BNNs are intertwined with their activation patterns based on the principle of ``\textit{Cells that fire together wire together}'', which is known as \textbf{cell assembly theory}. It explains and relates to several characteristics and advantages of BNN architecture such as modularity [\cite{peyrache2010principal}], efficiency, and generalizability, that are just the aspects in which the current DNNs are usually struggling [\cite{lecun2015deep}]. 
To take advantage of the beneficial architectural features in BNNs and overcome the existing drawbacks of DNNs, we propose the Biologically-Enhanced Artificial Neuronal assembly (BEAN) regularization. BEAN ensures neurons which “wire” together with a high outgoing weight correlation also “fire” together with small divergence in terms of their activation patterns.

An example of the artificial neuronal assembly achieved by our method can be seen in Figure \ref{fig:1}(d). The regularization is formulated as follows:
\begin{equation}
\small
\label{eq:consis_loss}
L^{(l)}_c = 1/ (S N_l^2)\sum\nolimits_{s}\sum\nolimits_{i}\sum\nolimits_{j} A^{(l)}_{i,j} \times d(H^{(l)}_{s,i},H^{(l)}_{s,j}) 
\end{equation}
where $L_c$ is the regularization loss; the term $A^{(l)}_{i,j}$ characterizes the wiring strength (the higher value, the stronger connection) between two neurons $i$ and $j$ within layer $l$; the term $d(H^{(l)}_{s,i},H^{(l)}_{s,j})$ models the divergence of firing patterns (the higher value, the more different the firing) between two neurons $i$ and $j$ on input sample $s$.
Thus, by multiplying these two functions, we  penalize those neurons with strong connectivity but high activation divergence, in line with the principles of cell assembly theory. $S$ is the total number of input samples while $N_l$ is the total number of hidden neurons in layer $l$.

Specifically, $A_{i,j}^{(l)}$ defines the connectivity relation among  neuron $i$ and neuron $j$ in DNN, which is instantiated by our newly proposed ``Layer-wise Neuron Correlation'' and will be elaborated in Sections \ref{sec:first_order} and \ref{sec:second_order}. On the other hand, to model the ``co-firing'' correlation, $d(H^{(l)}_{s,i},H^{(l)}_{s,j})$ is defined as ``Layer-wise Neuron Co-activation Divergence''  which denotes the difference in the activation patterns in $l$th layer between $H^{(l)}_{s,i}$ and $H^{(l)}_{s,j}$ of neuron $i$ and neuron $j$, respectively. Here $H^{(l)}_{s,i}$ represents the activation of neuron $i$ in layer $l$ for a given input sample $s$. The function $d(x,y)$ can be a common divergence metric such as absolute difference or square difference. In this study, we show the results for a square difference in the Experimental Study Section; the absolute difference results follow a similar trend.

\textbf{Model Training: }The general objective function of training a DNN model along with the proposed regularization on fully connected layer $l$ can be written as:
$L = L_{DNN} + \alpha L^{(l)}_c$
, where $L_{DNN}$ represents the general deep learning model training loss and the hyper-parameter $\alpha$ controls the relative strength of the regularization. 

Equation \ref{eq:consis_loss} can be optimized with backpropagation [\cite{rumelhart1988learning}] using the chain rule:
\begin{equation}
\small
\label{eq:bp}
\frac{\partial L^{(l)}_c}{\partial W^{(l+1)}} = \frac{\partial A^{(l)}}{\partial W^{(l+1)}} D^{(l)} , \
\frac{\partial L^{(l)}_c}{\partial W^{(l)}} = A^{(l)} \frac{\partial D^{(l)}}{\partial H^{(l)}} \frac{\partial H^{(l)}}{\partial W^{(l)}}, \ ...
\end{equation}
where $D^{(l)}\in\mathbb{R}^{S \times N_l \times N_l}$ of which each element is $D^{(l)}_{s,i,j}=d(H^{(l)}_{s,i},H^{(l)}_{s,j})$.
\begin{remark}
BEAN regularization has several strengths. 
First, it enforces interpretable neuronal assemblies without the need to introduce sophisticated handcrafted designs into the architecture, which is justified later in Section 3.1.
In addition, modeling the neuron correlations and dependencies further results in sparse and efficient connectivity in dense layers, which substantially reduced the computation/memory cost of the model, as shown in Section 3.2. 
Besides, the encoding of rich semantic correlation among neurons may improve the generalizability of the model when insufficient data and knowledge are provided, which is demonstrated later in Section 3.3.
Finally, the \textit{Layer-wise Neuron Correlation} can be efficiently computed with matrix operations, as per Equations \ref{eq:1_r_martix} and \ref{eq:2_r_martix}, which enables modern GPUs to boost up the speed during model training. In practice, we observe negligible run time overhead of the addition computation needed for BEAN regularization.
\end{remark}

\subsection{The First-Order Layer-wise Neuron Correlation}
\label{sec:first_order}

\begin{figure}
\centering
\includegraphics[width=\linewidth]{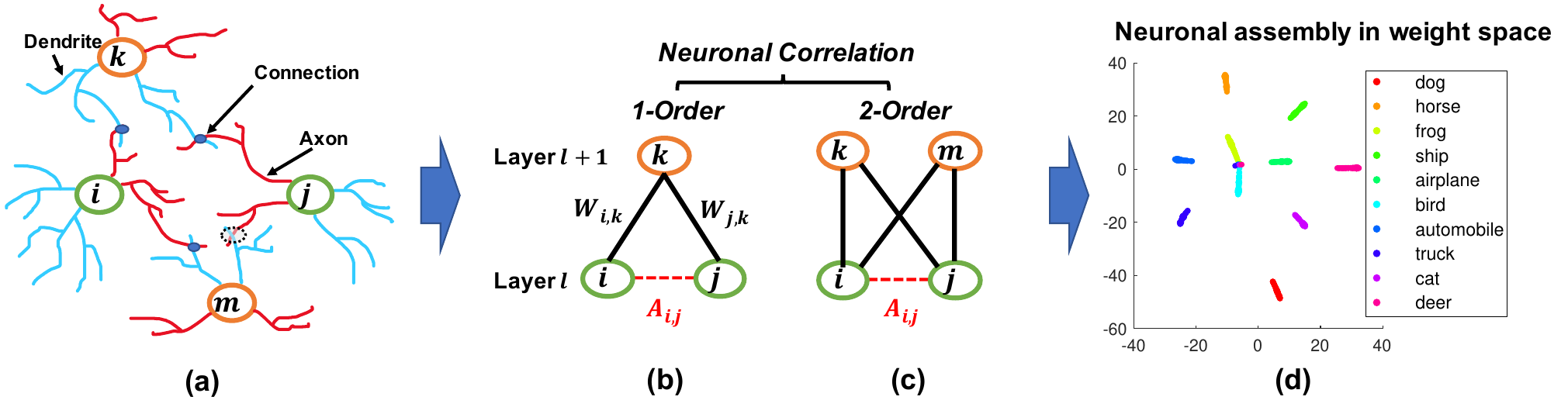}
\caption{An illustration of how the proposed constraint drew inspiration from BNNs and bipartite graphs. \textbf{(a)} neuron correlations in BNNs correspond to connections between dendrites, which are represented by blue lines, and axons, which are represented by red lines. \textbf{(b)} and \textbf{(c)} analogy of figure (a) represented as connections between layers in DNNs; although nodes $i$ and $j$ cannot form direct links, they can be correlated by a given node $k$ as a first-order correlation, or by two nodes $k$ and $m$ as a second-order correlation which is also equivalent to a 4-cycle in bipartite graphs. \textbf{(d)} an example of a learned neuronal assembly in neurons outgoing weight space, with the dimensionality reduced to 2D with T-SNE [\cite{maaten2008visualizing}]. Each point represents one neuron and the neurons are colored according to their highest activated class in the test data.}
\label{fig:1}
\end{figure}

This section introduces the formulation of the layer-wise neuron correlation  $A_{i,j}^{(l)}$ between any pair of neurons $i$ and $j$.

In the human brain, the correlation between two neurons depends on the wiring between them [\cite{buzsaki2010neural}] and hence is typically treated as a binary value in BNN studies, with ``1'' indicating the presence of a connection and ``0'' the absence, so the correlation among a group of neurons can be represented by the corresponding adjacency matrix. 
Although there is typically no direct connection between neurons within the same layer of DNNs, it is possible to model neuron correlations based on their connectivity patterns to the next layer.
This resembles a common approach in network science, where it is useful to consider the relationships between nodes based on their common neighbors in addition to their direct connections.
One classic concept widely used to describe such a pattern is called \textit{triadic closure} [\cite{granovetter1977strength}]. As shown in Figure \ref{fig:1} (b), triadic closure can be interpreted here as a property among three nodes $i$, $j$, and $k$, such that if connections exist between $i-k$ and $j-k$, there is also a connection between $i-j$. 

We take this scheme a step further to model the correlations between neurons within the same layer by their connections to the neurons in the next layer. This can be considered loosely analogous to the degree of similarity of the axonal connection pattern of biological neurons in BNNs [\cite{rees2017weighing}].
To simulate the relative strength of such connections in DNNs, we introduce a function $f(\cdot)$ that converts the actual weights into a relative connectivity strength.
Suppose matrix $W^{(l+1)}\in\mathbb{R}^{N_l \times N_{l+1}}$ represents all the weights between neurons in layers $l$ and $l+1$ in DNNs, where $N_l$ and $N_{l+1}$ represent the numbers of neurons, respectively. 
The relative connectivity strength can be estimated by the following equation\footnote{Similar to the ReLU activation function, our formulation introduces a non-differentiable point at zero; we follow the conventional setting by using the sub-gradient for model optimization.}:
\begin{equation}
\small
\label{eq:W_hat}
f(W^{(l+1)})=|tanh(\gamma W^{(l+1)})|
\end{equation}
where $|\cdot|$ represents the element-wise absolute operator; $tanh(\cdot)$ represents the element-wise hyperbolic tangent function; and $\gamma$ is a scalar that controls the curvature of the hyperbolic tangent function.
The values of $f(W^{(l+1)})\in\mathbb{R}^{N_l \times N_{l+1}}$ will all be positive and in the range of $[0,1)$ with the value simulating the relative connectivity strength of the corresponding synapse between neurons. 

Although there can be positive and negative weights in DNNs, our assumption on connection strength follows the typical way of BNN studies, which measures the presence and absence of the connection as mentioned above. 
Moreover, since DNNs require continuous values instead of discrete values to make the function differentiable for optimization, we further use Equation \eqref{eq:W_hat} to convert the concept of the presence/absence of the connections to the relative strength of the connections.  
More specifically, the difference is that instead of treating connection to be either ``1'' (indicating the presence of a connection) or ``0'' (indicating the absence of the connection), we treat the output of Equation \eqref{eq:W_hat} as the strength of that connection, where high values (i.e. close to ``1'') indicate the presence of a strong connection and low values (i.e. close to ``0'') indicate weak or no connection.

Based on this, we can now give the definition for the \textit{layer-wise first-order neuron correlation} as:
\begin{definition}{\textbf{Layer-wise first-order neuron correlation.}}
\label{the:1}
For a given neuron $i$ and neuron $j$ in layer $l$, the layer-wise first-order neuron correlation is given by:
\begin{equation}
\small
\label{eq:1_r_element_wise}
A^{(l)}_{i,j}=(1/N_{l+1})\sum\nolimits_{k=1}^{N_{l+1}} f(W^{(l+1)}_{i,k}) \times f(W^{(l+1)}_{j,k})
\end{equation}
The above formula can be expressed as the product of two matrices:
\begin{equation}
\small
\label{eq:1_r_martix}
A^{(l)} = (1/N_{l+1})f(W^{(l+1)})\cdot f(W^{(l+1)})^T
\end{equation}
where $\cdot$ represents the matrix multiplication operator.
\end{definition}

The layer-wise neuron correlation matrix  $A^{(l)}$ is a symmetric square matrix that models all the pairwise correlations of neurons with respect to their corresponding outgoing weights in layer $l$.
Each entry $A^{(l)}_{i,j}$ takes a value in the range $[0,1)$ and models the correlation between neuron $i$ and neuron $j$ in terms of the similarity of their connectivity patterns. The higher the value, the stronger the correlation between the two.

In this setting, two neurons $i$ and $j$ from layer $l$ will be linked and correlated by an intermediate node $k$ from layer $l+1$ if and only if both edges $f(W^{(l+1)}_{i,k})$ and $f(W^{(l+1)}_{j,k})$ are non zero, and the relative strength can be estimated by $f(W^{(l+1)}_{i,k}) \times f(W^{(l+1)}_{j,k})$, which will be in the range $[0,1)$. Since there are $N_{l+1}$ neurons in layer $l+1$, where each neuron $k$ can contribute to such connections, running over all neurons in layer $l+1$ we obtain Equation \ref{eq:1_r_element_wise} and Equation \ref{eq:1_r_martix}.

\subsection{The Second-Order Layer-wise Neuron Correlation}
\label{sec:second_order}

Although the first-order correlation is able to estimate the degree of dependency between each pair of neurons, it may not be sufficient to strictly reflect the degree of grouping or assembly of the neurons. 
Thus, here we further propose a second-order neuron correlation based on the first-order correlation defined in Equation \ref{eq:1_r_element_wise} and \ref{eq:1_r_martix}, as:
\begin{definition}{\textbf{Layer-wise second-order neuron correlation.}}
For a given neuron $i$ and neuron $j$ in layer $l$, the layer-wise second-order neuron correlation is given by:
\begin{equation}
\small
\label{eq:2_r_element_wise}
A^{(l)}_{i,j}=(1/N_{l+1}^2)\sum\nolimits_{k,m} f({W}^{(l+1)}_{i,k}) \times f({W}^{(l+1)}_{j,k}) \times f({W}^{(l+1)}_{i,m}) \times f({W}^{(l+1)}_{j,m})
\end{equation}
The above formula can be expressed as the product of four matrices:
\begin{equation}
\small
\label{eq:2_r_martix}
A^{(l)} = (1/N_{l+1}^2)(f({W}^{(l+1)})\cdot f({W}^{(l+1)})^T)\odot(f({W}^{(l+1)})\cdot f({W}^{(l+1)})^T)
\end{equation}
where $\odot$ represents the element-wise multiplication of matrices.
\end{definition}
The second-order correlation provides a stricter criterion for relating neurons, as it requires at least two common neighbor nodes from the layer above to have strong connectivity, as compared to the first-order correlation that requires just one common neighbor. 
Moreover, the second-order neuron correlation is closely related both to graph theory concepts and a  neuroscience-inspired learning rule:

\begin{remark}{\textbf{Graph theory and neuroscience interpretation.}}
Modeling the first-order correlation between two neurons within the same layer is based on the co-connection to a common neighbor neuron from the layer above, which is closely related to the concepts of \textit{clustering coefficient} [\cite{watts1998collective}] and \textit{transitivity} [\cite{holland1971transitivity}] in graph theory. On the other hand, modeling the second-order correlation between two neurons involves two common neighbor neurons in the layer above, which is closely related to calculating the 4-cycle pattern where all 4 possible connections in between are taken into account, as shown in Figure \ref{fig:1} (b). 
This 4-cycle pattern is linked to the global clustering coefficients of bipartite networks [\cite{robins2004small}], where the set of vertices can be decomposed into two disjoint sets such that no two vertices within the same set are adjacent. Similarly, if we consider neurons within one layer as the nodes that belong to one set of the bipartite network between two adjacent layers of the neural networks, forming this 4-cycle will tend to increase the clustering coefficients of the network.
Moreover, the second-order correlation is also related to several cognitive neuroscience studies, such as the BIG-ADO learning rule and the principal semantic components of language [\cite{mainetti2015neural, samsonovich2010toward}] as well as the notion of discrete neuronal circuits [\cite{pulvermuller2009discrete}].
Figure \ref{fig:1} (a) illustrates a scenario of the BIG-ADO learning rule in BNNs. The blue blobs represents a connection that was formed between two neurons (i.e., a synapse), while the dashed circle between neurons $j$ and $m$ represents an Axo-Dendritic Overlap (ADO) (i.e., a potential synapse) between the two neurons. 
BIG-ADO posits that in order to form a synapse, there must be a potential synapse in place, and the probability of having a potential synapse grows with the second-order correlation.
Notably, both of the neuroscience papers cited above
relate such a learning mechanism to
the formation of cell assemblies in the brain, which parallels our observation of neuronal functional clusters among neurons in DNNs when BEAN was imposed, as shown in Figure \ref{fig:1} (c) and Figure \ref{fig:few_shot_study} (b).
\end{remark}

\section{Experimental Study}
Our description of the empirical analysis design and results is organized in the following fashion. In Section 3.1, we first characterize the interpretable patterns from the learning outcomes of BEAN regularization on multiple classic image recognition tasks. 
We then further analyze in Section 3.2 how BEAN could benefit the model from learning sparse and efficient neuron connections.
Finally, in Section 3.3 we study the effect of BEAN regularization on improving the generalizability of the model on several few-shot learning from scratch task simulations. 
We refer to both distinct BEAN variations, BEAN-1 and BEAN-2, based on the two proposed layer-wise neuron correlation defined by Equation \ref{eq:1_r_martix}, and Equation \ref{eq:2_r_martix} respectively. 
The value for $\gamma$ (Equation \ref{eq:W_hat}) was set to 1.
This paper focuses on examining the effects of the proposed regularization rather than the differences between distinct types of neural network architectures. Hence, we simply adopted several of the most popular neural network architectures for the chosen datasets and did not perform any hyperparameter or system parameter tuning using the test set; in other words, 
we did not perform any "post selection" (i.e. selectively reporting the model results based on testing set [\cite{zheng2016challenges, zheng2016mobile}]). 
All network architectures used in this paper are fully described in their respective cited references, including the specification of their system parameters. The regularization factor of BEAN and other baseline methods were chosen based on the model performance on the validation set.
All the experiments were conducted on a 64-bit machine with Intel(R) Xeon(R) W-2155 CPU \@3.30GHz processor and 32GB memory and an NVIDIA TITAN Xp GPU.

\subsection{The Interpretable Patterns of BEAN Regularization}

Due to the highly complex computation among numerous layers of neurons in traditional DNNs, it is typically difficult to understand how the network learned what it remembers and the system is more commonly treated as a black-box model [\cite{zhang2018interpretable}]. 
Here, to ascertain the effect of BEAN regularization on the interpretability of network dynamics, we analyze the differences in neuronal representation properties of the DNNs with and without BEAN regularization.
We conducted experiments on three classic image recognition tasks on the MNIST [\cite{lecun1998gradient}], Fashion-MNIST [\cite{xiao2017fashion}] and CIFAR-10 [\cite{krizhevsky2009learning}] datasets by starting with three predefined network architectures as listed below:
\begin{enumerate}
\item An MLP with one hidden layer of 500 neurons with ReLU activation function for MNIST and Fashion-MNIST datasets.
\item A LeNet-5 [\cite{lecun1998gradient}] for MNIST and Fashion-MNIST datasets.
\item ResNet18 [\cite{he2016deep}] for CIFAR-10 dataset.
\end{enumerate}
The Adam optimizer [\cite{kingma2014adam}] was used with a learning rate of 0.0005 and a batch size of 100 for model training until train loss convergence was achieved; BEAN was applied to all the dense layers of each model. 

\subsubsection{Biological plausibility of the learned neuronal assemblies}
\begin{figure}
\centering
\includegraphics[width=\linewidth]{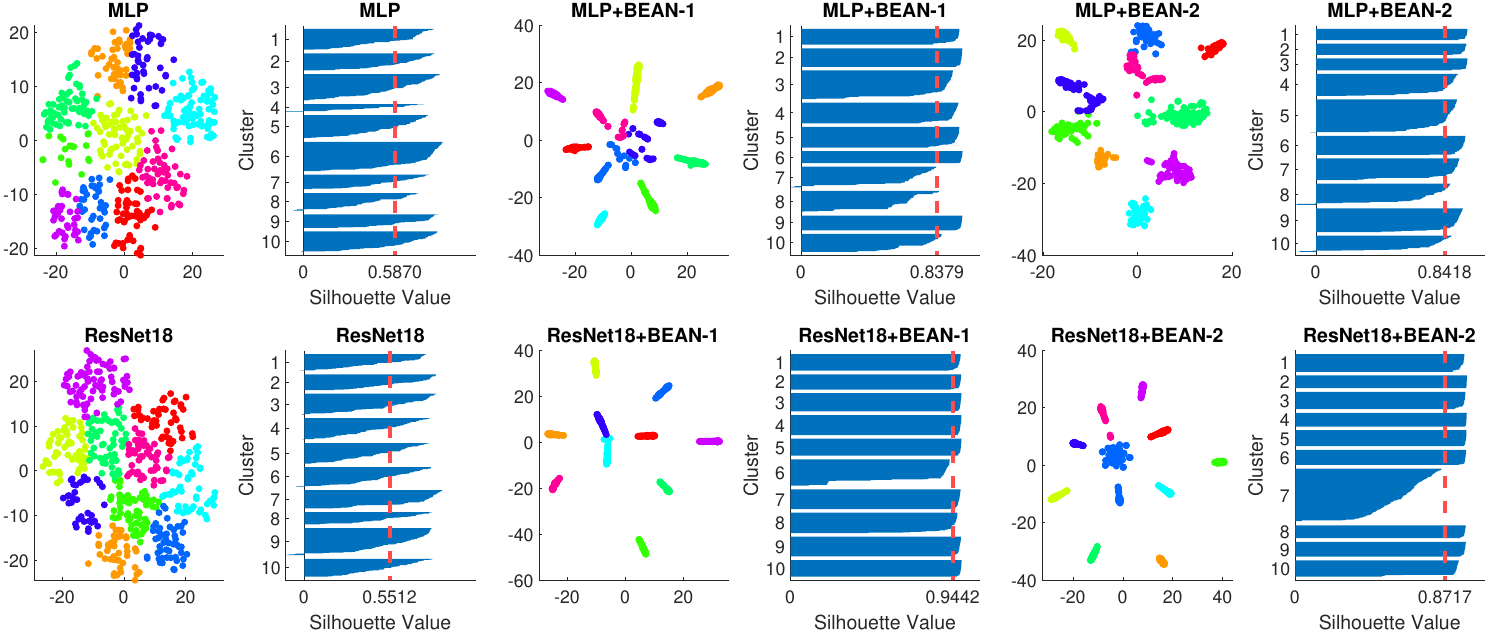}
\caption{Neuronal assembly patterns found in neurons' weight space of the dense layer of different models on both MNIST (top) and CIFAR-10 (bottom) datasets, along with clustering validation via Silhouette score on 10 clusters K-means clustering.
The dimensionality of neurons' weight space was reduced to 2D with T-SNE for visualization.}
\label{fig:sihouette}
\end{figure}

By analyzing the neurons' connectivity patterns based on their outgoing weights, we discovered neuronal assemblies in dense layers where BEAN regularization was enforced. Specifically, for both datasets, we found that the neuronal assemblies at the last dense layer could be best described by 10 clusters with K-means clustering [\cite{macqueen1967some}] validated by Silhouette analysis [\cite{rousseeuw1987silhouettes}]. Silhouette analysis is a widely-used method for interpretation and validation of consistency within clusters of data. The technique provides a succinct graphical representation of how well each object has been classified. As shown in Figure \ref{fig:sihouette}, we visualized the K-means clustering results in neurons' weight space of the dense layer on both MNIST (top) and CIFAR-10 (bottom) datasets. Each data point in the figure indicates one single neuron and the color indicates its cluster assignment by the clustering algorithm. The Silhouette value is further used to assess the quality of the clustering assignment: high Silhouette values support the existence of clear clusters in the data points, which here correspond to neural assembly patterns among neurons.

Both BEAN-1 and BEAN-2 could enforce neuronal assemblies for various models on several datasets, yielding Silhouette indices around 0.9, which indicates strong clustering patterns among neurons in dense layers where BEAN regularization was applied. On the other hand, training conventional DNN models with the same architectures could only yield Silhouette indices near 0.5, which indicates no clear clustering patterns in conventional dense layers of deep neuronal networks.

\begin{figure}
\centering
\includegraphics[width=\linewidth]{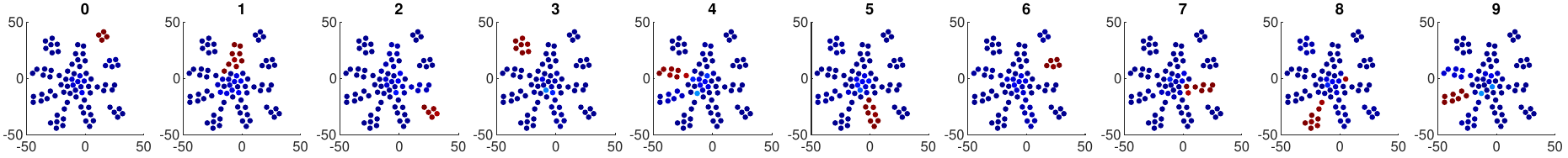}
\caption{Neuron co-activation patterns found in the representation of the last dense layer of LeNet-5+BEAN-2 model on MNIST dataset. 
The dimensionality of neurons' weight space was reduced to 2D with T-SNE for visualization.
Each point represents one neuron within the last dense layer of the model and is colored based on its activation scale. The 10 subplots show the average activation heat-maps when each digit's samples were fed into the model. The warmer color indicates a higher neuron activation.}
\label{fig:activation_cnn}
\end{figure}

Moreover, we found co-activation behavior of neurons within each neuronal assembly, which is both interpretable and biologically plausible. Figure \ref{fig:activation_cnn} shows the visualization of neuron co-activation patterns found in the last dense layer of LeNet-5+BEAN-2 model on MNIST dataset. 
For the samples of each specific class, only those neurons in the specific neuron group that is associated with that digit class have high activation while all the other neurons remain silent. This strong correlation between each unique assembly and each unique class concept allows straightforward  interpretation of the neuron populations in the dense layers. 
From the neuroscience perspective, those co-activation patterns and the association between high-level concepts and neuron groups may reflect similar co-firing patterns observed in biological neural systems  [\cite{peyrache2010principal}] and underscore the strong association between neuronal assembly and concepts [\cite{tononi2003measuring}] in biological neural networks.

We also found a strong correlation between neuronal assembly and class selectivity indices. Selectivity index was originally proposed and used in systems neuroscience [\cite{de1982orientation, freedman2006experience}]. Recently, machine learning researchers also studied unit class selectivity  [\cite{morcos2018importance, zhou2018revisiting}] as a metric for interpreting the behaviors of single units in deep neural networks. Mathematically, it is calculated as:
$
selectivity = (\mu_{max}-\mu_{-max}) / (\mu_{max}+\mu_{-max})
$
, where $\mu_{max}$ represents the highest class-conditional mean activity and $\mu_{-max}$ represents the mean activity across all other classes. 

\begin{figure}
\centering
\includegraphics[width=\linewidth]{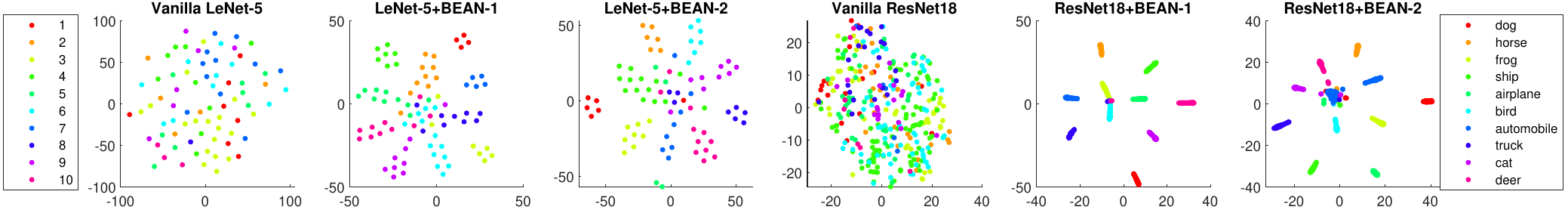}
\caption{The strong association between neuronal assemblies and neurons' class selectivity index with BEAN regularization on both MNIST (left) and CIFAR-10 (right) datasets. Each point represents one neuron and the color represents the class where the neuron achieved its highest class-conditional mean activity in the test data.}
\label{fig:selectivity_cnn_resnet}
\end{figure}

To better visualize how high-level concepts are associated with the learned neuron assemblies, we further labeled each neuron with the class in which it achieved its highest class-conditional mean activity $\mu_{max}$ in the test data. 
Figure \ref{fig:selectivity_cnn_resnet} shows the results for the last dense layer of the models trained with both datasets. We found that the neuronal assembly could be well described based on selectivity. The strong association between neuronal assemblies and neurons' selectivity index further demonstrated the biological plausibility of the learning outcomes of BEAN regularization. 
Moreover, the strong neuron activation patterns towards each individual high-level concepts or classes could in principle enable one to better understand what each individual neuron has learned to represent. However, more relevant to and consistent with our regularization, these selective activation patterns reveal how a group of neurons (i.e. neuronal assembly) together capture the whole picture of each high-level concept, such as the `bird' class in CIFAR-10 as shown in Figure \ref{fig:selectivity_cnn_resnet}.

In this subsection, we have demonstrated the promising effect of the proposed BEAN regularization on forming the neural assembly patterns among the neurons in the last layer of the network and their correspondence with biological neural networks. Although the effect of BEAN regularization is not yet clear on the lower layers of the networks, it will be interesting in the future to explore additional relations between computational function and the architecture of earlier processing stations in biological neural systems.

\subsubsection{Quantitative analysis of interpretability}
Experimental neuropsychologists commonly use an ablation protocol when studying neural function, whereas parts of the brain are removed to investigate the cognitive effects. Similar ablation studies have also been adapted for interpreting deep neural networks, such as understanding which layers or units are critical for model performance [\cite{girshick2014rich,morcos2018importance,zhou2018revisiting}].

\begin{figure}
\centering
\includegraphics[width=\linewidth]{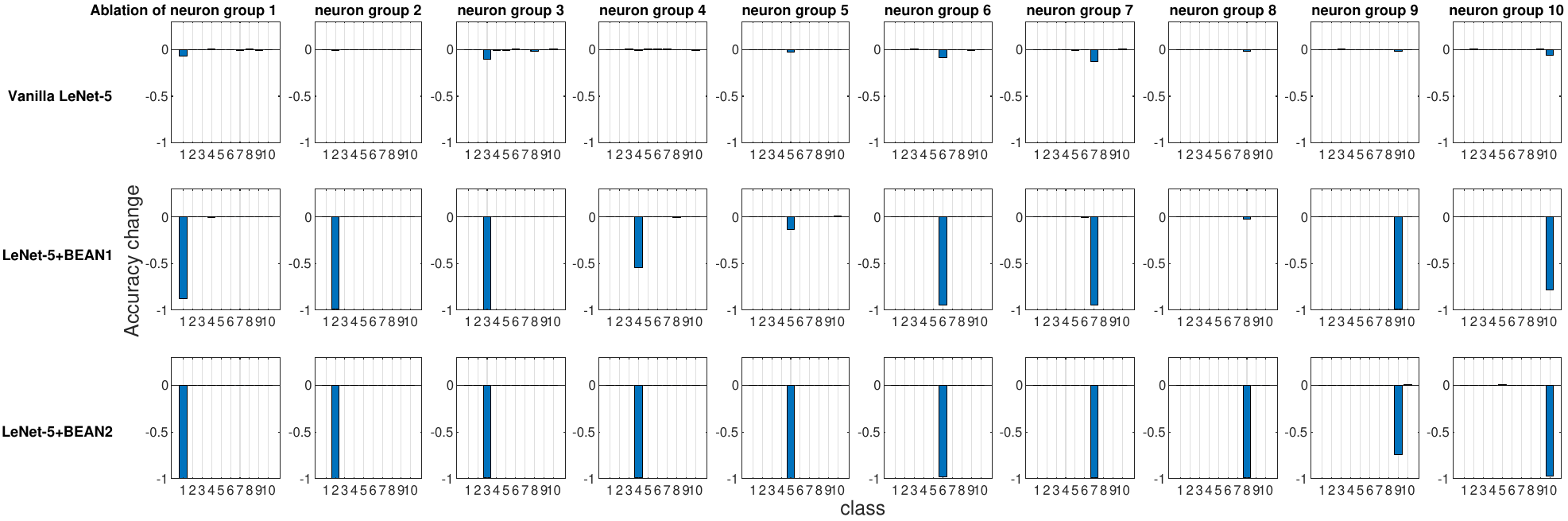}
\caption{The ablation study at the neuron population level of the last dense layer of LeNet-5 models. Each time, one distinct group of neurons were ablated based on their most selective class and the model performance changes for each individual class were recorded.}
\label{fig:ablation_cnn}
\end{figure} 

To quantitatively evaluate and compare interpretability, we performed an ablation study at the neuron population level, each time ablating one distinct group of neurons and recording the consequent model performance changes for each class. As shown in Figure \ref{fig:selectivity_cnn_resnet}, we identified neuron groups via class selectivity and performed neuron population ablation accordingly.
Figure \ref{fig:ablation_cnn} shows the results of all 10 ablation runs for each class in MNIST dataset.
As also reported by [\cite{morcos2018importance}], for conventional deep neural nets, there is indeed no clear association between neuron's selectivity and importance to the overall model performance, as revealed by neuron population ablation. However, when BEAN regularization was utilized during training, such association clearly emerged, especially for BEAN-2.
This is because BEAN-2 could enforce neurons to form stricter neuron correlations than BEAN-1 with the second-order correlation, enabling groups of neurons to represent more compact and disentangled concepts, such as handwritten digits. 
This discovery further demonstrated the interpretability and concept level representation in each neuronal assembly learned by applying BEAN regularization.
Such compact and interpretable structure of concept-level information encoding could also benefit the field of disentanglement representation learning [\cite{bengio2013representation}].

\subsection{Learning Sparse and Efficient Networks}
\label{sec:3.2}

To evaluate the effect of BEAN regularization on learning sparse and efficient networks, we conducted experiments on two real-world benchmark datasets, i.e., the MNIST [\cite{lecun1998gradient}] and Fashion-MNIST [\cite{xiao2017fashion}] datasets. We compared BEAN with several state-of-the-art regularization methods that could enforce sparse connection of the network, including $\ell_{1}$-norm, group sparsity based on $\ell_{2,1}$-norm [\cite{yuan2006model,alvarez2016learning}], and exclusive sparsity based on $\ell_{1,2}$-norm [\cite{zhou2010exclusive,kong2014exclusive}].
Notable studies also investigated the combination of the sparsity terms listed above, such as combining group sparsity and $\ell_{1}$-norm [\cite{scardapane2017group}], and combining group and exclusive sparsity [\cite{yoon2017combined}].
The combinatorial study is outside the scope of this work, as our focus is on showing and comparing the effectiveness of the single regularization term to the network.
To keep the comparison fair and accurate, we use the same base network architecture for all regularization methods tested in this experiment, which is a predefined fully connected neural network with 3 hidden layers, 500 neurons per layer, and ReLU as the neuron activation function. 
The regularization methods are applied to all layers of the network, except the bias term. 
The regularization co-efficients are selected through a grid search varying from $10^{-5}$ to $10^{3}$ based on the model performance on the validation set, as shown in Algorithm 1.
To obtain a more reliable and fair result, we ran a total of 20 random weight initializations for every network architecture studied and reported the overall average performance of all 20 results as the final model performance of each architecture.

\begin{algorithm}
\small
\caption{The pseudo code for searching for the best $\alpha$ value in BEAN}
    $\text{func hyperparameter\_tuner}(\text{training\_data, validation\_data, alpha\_list} =[0.001, 0.01, 0.1, 1, 10, 100]):$\\
        \Indp 
        $\text{hp\_perf} = []$ \\
        \textit{\% train and evaluate on all hyper-parameter settings}\\
        \text{foreach} $\alpha \text{ in } \text{alpha\_list}:$ \\
            \Indp
            $m = \text{train\_model} (\text{training\_data}, \text{alpha})$\\
            $\text{validation\_results} = \text{eval\_model}(m, \text{validation\_data})$\\
            $\text{hp\_perf.}\text{append}(\text{validation\_results})$\\
            \Indm
  \textit{\% find the best alpha on validation set}\\
  $\text{best\_alpha} = \text{alpha\_list}[\text{max\_index}(\text{hp\_perf})]$\\
  \text{return} \text{best\_alpha}
\end{algorithm}

To quantitatively measure the performance of various sparse regularization techniques, we used three evaluation metrics, including the prediction accuracy on test data (i.e. measured by the number of correct predictions divided by the total number of samples in test data), the ratio of parameters used in the network (i.e. total number of non-zero weights divided by the total number of weights in the networks after training), and the corresponding number of floating point operations (FLOPs). A higher accuracy means that the model can train a better network for the classification tasks. 
A lower FLOP indicates that the network needs fewer computational operations per forwarding pass, which reflects computation efficiency. 
Similarly, a lower parameter usage indicates the network requires less memory usage, which reflects memory efficiency. 

\begin{table*}[ht]
  \caption{Efficient model learning experiments on MNIST and Fashion-MNIST datasets. 
  The FLOPs and effective parameters (i.e. number of non-zero parameters) are normalized by the value of vanilla model.
  Performance is averaged over 20 runs.
  The best and second-best results are highlighted in boldface and italic font, respectively.}
  \centering
  \label{tab:efficient_learning}
  \resizebox{\textwidth}{!}{
  \begin{tabular}{c|c|cccc|cc}
    \toprule
    Dataset & Measure   & Vanilla   & $\ell_{1}$-norm    & Group Sparsity& Exclusive Sparsity& BEAN-1        & BEAN-2         \\
    \hline
    \multirow{3}{*}{MNIST}&
            accuracy    & 0.9812    &\textit{0.9835}& 0.9813        & 0.9824            &\textbf{0.9842}& 0.9823         \\
        &   FLOPs        & 1         & 0.8106        & 0.6098        & 0.4248            &\textit{0.2212}& \textbf{0.1320}\\
        &   parameter   & 1         &0.2921         &\textit{0.0982}& 0.1375            & 0.1496        & \textbf{0.0730}\\
    \hline
    \multirow{3}{*}{Fashion-MNIST}&
            accuracy    & \textbf{0.8986}&0.8924    & 0.8925        & 0.8930            &\textit{0.8960}& 0.8916         \\
        &   FLOPs        & 1         & 0.8011        & 0.5384        & 0.5320            &\textit{0.2913}& \textbf{0.1622}\\
        &   parameter   & 1         &0.4357         &\textit{0.1378}& 0.2257            & 0.2592        & \textbf{0.1259}\\
    \bottomrule
  \end{tabular}}
\end{table*}

The results are shown in Table \ref{tab:efficient_learning}. For each evaluation metric, the best and second-best results are highlighted in boldface and italic font, respectively. As can be seen, both BEAN-1 and BEAN-2 can achieve high memory and computational efficiency without sacrificing network performance for the classification tasks.
Specifically, BEAN-2 achieved the best memory and computational efficiency, out-performing baseline models by 25-75\% on memory efficiency and 69-84\% on computational efficiency on the MNIST dataset, and by 9-71\% on memory efficiency and 69-80\% on computational efficiency on the Fashion-MNIST dataset. 
BEAN-1 also achieved a good trade-off between model performance and efficiency, being the second-best on computational efficiency and the best on model performance on both the MNIST and Fashion-MNIST datasets. Comparing with BEAN-2, BEAN-1 leans more toward the model performance side in such a trade-off. This is because the first-order correlation used in BEAN-1 is less restrictive than a higher-order correlation in BEAN-2, as only one support neuron in the layer above is enough to build up a strong correlation. Thus, in practice, using a higher-order correlation might be promising when the objective is to learn a more efficient model. 

Interestingly, BEAN regularization seemed to advance the state-of-the-art by an even more significant margin in terms of computational efficiency.
In fact, BEAN regularization reduces the number of FLOPs needed for the network by automatically "pruning" a substantial proportion of neurons in the hidden layers (whereas a neuron is considered pruned if either all incoming or all outgoing weights are zero), due to the penalization of connections between neurons that encode divergent information. 
Although group sparsity and exclusive sparsity are designed to achieve a similar objective for obtaining neuron-level sparsity, they are less effective than BEAN regularization. This is due to the fact that BEAN takes into consideration not only the correlations between neurons via their connection patterns but also the consistency of those correlations with their activation patterns.

We have shown in Table \ref{tab:efficient_learning} that the proposed BEAN regularization can effectively make the connection sparser in the dense layers of the artificial neural networks. In general, this ‘sparsifying’ effect can be beneficial for any models with at least one dense layer in the network architecture. Most modern deep neural networks (such as VGG [\cite{simonyan2014very}] and ImageNet [\cite{russakovsky2015imagenet}]) can enjoy this sparsity benefit, as the dense layers typically contribute to the majority of the model parameters [\cite{cheng2015exploration}].

\subsection{Towards few-shot learning from scratch with BEAN regularization}
In an attempt to test the influence of BEAN regularization on the generalizability of DNNs in the scenarios where the training samples are extremely limited, we conducted a \textit{few-shot learning from scratch} task, i.e. without the help of any additional side tasks and pre-trained models [\cite{kimura2018few}]. 
Notice that in the few-shot learning setting, the model typically requires an iterative learning process over the sample set. In other words, for each individual few-shot learning experiment, only a few image samples per digit are randomly selected to form the training set. The model then iteratively learns from the selected image samples until convergence is achieved.
So far, this kind of learning task has rarely been explored due to the difficulty of the problem setup as compared to other conventional few-shot learning tasks where additional data or knowledge could be accessed. 
Currently, only [\cite{kimura2018few}] carried out a preliminary exploration with their proposed Imitation Networks model. 
We conducted several simulations of the \textit{few-shot learning from scratch} task on the MNIST [\cite{lecun1998gradient}], Fashion-MNIST [\cite{xiao2017fashion}], and CIFAR-10 [\cite{krizhevsky2009learning}] datasets. 
Besides Kimura's Imitation Networks, we also compared BEAN with other conventional regularization techniques commonly used in the deep learning literature. Specifically, we compared dropout [\cite{srivastava2014dropout}], weight decay [\cite{krogh1992simple}], and $\ell_{1}$-norm.
Similarly to the description of Section \ref{sec:3.2}, we kept the comparison fair and accurate by using a predefined network architecture, namely LeNet-5 [\cite{lecun1998gradient}], as the base network architecture for all regularization methods studied in this experiment. The regularization terms were applied to all three dense layers of the base LeNet-5 network. Once again, the hyperparameter of each regularization along with all other system parameters were selected through a grid search and based on the best performance on a predefined 10k validation set sampled from the original training base and completely distinct from the training samples used in the few-shot learning tasks and the testing set.

\begin{table}
  \caption{Few-shot learning from scratch experiments on the MNIST (left), Fashion-MNIST (middle), and CIFAR-10 (right) datasets. Performance is averaged over 20 simulations of randomly sampled training data from the original training base. The best and second-best results for each few-shot learning setting are highlighted in boldface and italic font, respectively. }
  \label{tab:few-shot-MNIST}
  \centering
  \resizebox{\textwidth}{!}{
  \begin{tabular}{c|cccc|cccc|cccc}
    \toprule
    Dataset         &\multicolumn{4}{c|}{MNIST}      &\multicolumn{4}{c}{Fashion-MNIST}     &\multicolumn{4}{|c}{CIFAR-10}\\
    \hline
    Model           &1-shot     &5-shot     &10-shot    &20-shot
                    &1-shot     &5-shot     &10-shot    &20-shot
                    &1-shot     &5-shot     &10-shot    &20-shot\\
    \hline
    Vanilla            &38.63      &70.21      &78.97      &86.68
                                &39.32      &59.02      &64.50      &70.23
                                &15.60      &18.49      &22.45      &26.39\\
    \hline
    Dropout           &40.13      &72.45      &82.04      &89.22
                                &40.78      &60.04      &65.40      &71.83
                                &15.10      &18.85      &22.73      &26.01\\
    Weight decay      &39.51      &71.76      &82.87      &90.15
                                &41.31      &61.98      &67.25      &71.88
                                &15.47      &19.17      &23.74      &26.77\\
    $\ell_{1}$-norm   &40.96      &74.35      &81.17      &90.68
                                &41.26      &62.18      &67.30      &70.85
                                &15.64      &18.95      &23.16      &26.99\\
    Imitation networks          &44.10      &70.40      &80.00      &86.70
                                &44.80      &62.10      &68.00      &72.50
                                &\multicolumn{4}{c}{-}\\
    \hline
    BEAN-cos            &\textit{54.05}      &80.16      &86.28      &92.22
                                &42.48      &65.49      &68.97      &74.20
                                &18.23      &\textit{21.45}      &24.66      &28.74\\
    BEAN-1            &\textbf{54.79} &\textbf{83.42} &\textit{87.51} &\textit{92.79}
                                &\textbf{50.57} &\textbf{66.95} &\textit{69.21} &\textit{74.25}
                                &\textbf{19.39} &\textbf{21.92} &\textit{24.81} &\textit{28.95}\\
    BEAN-2            &53.75 &\textit{80.76} &\textbf{88.08} &\textbf{92.97}
                                &\textit{49.94} &\textit{65.98} &\textbf{70.21} &\textbf{75.06}
                                &\textit{19.28} &21.28 &\textbf{25.04} &\textbf{29.23}\\
    \bottomrule
  \end{tabular}}
\end{table}

Table \ref{tab:few-shot-MNIST} shows model performance on several \textit{few-shot learning from scratch} experiments on the MNIST, Fashion-MNIST, and CIFAR-10 datasets. Performance is averaged over 20 experiments of randomly sampled training data from the original training base. The best and second-best results for each few-shot learning settings are highlighted in boldface and italic font, respectively.
As can be seen, the proposed BEAN regularization advanced the state-of-the-art by a significant margin on all four \textit{few-shot learning from scratch} tasks tested among all three datasets.
Moreover, BEAN advanced the performance more significantly when training samples were more limited.
For instance, BEAN outperformed all comparison methods by 24-42\%, 13-29\%, and 24-28\% on 1-shot learning tasks on the MNIST, Fashion-MNIST, and CIFAR-10 datasets, respectively. This observation demonstrates the promising effect of BEAN regularization on improving the generalizability of the neural nets when the training samples are extremely limited.
Another interesting observation is that BEAN-1 in general performed the best with extremely limited training samples, such as the 1-shot and 5-shot learning tasks, while BEAN-2 regularization in general performed the best with slightly more training samples, such as the 10-shot and 20-shot learning tasks. 
The reason behind this observation might be related to the more stringent higher-order correlation, which requires more common neighbor neurons that appear to have strong connections with both neurons. Thus, a modestly increased availability of sample observations could enable BEAN-2 to  form more effective neuronal assemblies,  further improving the model performance.

Furthermore, we studied an additional variant for BEAN, i.e. BEAN-cos, which calculates the layer-wise neuron correlation via cosine similarity between the downstream weights of two neurons. As shown in Table \ref{tab:few-shot-MNIST}, we found that BEAN-cos can still yield good performance and beat other existing regularization methods, and getting competitive results as compared with BEAN-1 and BEAN-2. However,  it is inferior to BEAN-1 in 1-shot and 5-shot settings, and inferior to BEAN-2 in 10-shot and 20-shot settings. 
This is because BEAN-cos is unable to handle the order of correlation between neurons, as using cosine similarity requires us to treat the out-going weights of a neuron as a whole (vector) to compute the pair-wise similarity between neurons. Thus, doing this will lose the ability to calculate higher-order correlation (such as the second-order correlation), and consequentially lose the good interpretation from graph theory and neuroscience (as described in Remark 2).

To better understand why BEAN regularization could help the seemingly over-parameterized model generalize well on a small sample set, we further analyzed the learned hidden representation of the dense layers where BEAN regularization was employed. 
We found that BEAN helped the model gain better generalization power in two aspects: 1) by automatic sparse and structured connectivity learning and 2) by weak parameter sharing among neurons within each neuronal assembly. Both aspects enhanced the dense layers to promote efficient and parsimonious connections, which consequently prevented the model from over-fitting with a small training sample size.

\begin{figure}
\centering
\includegraphics[width=0.9\linewidth]{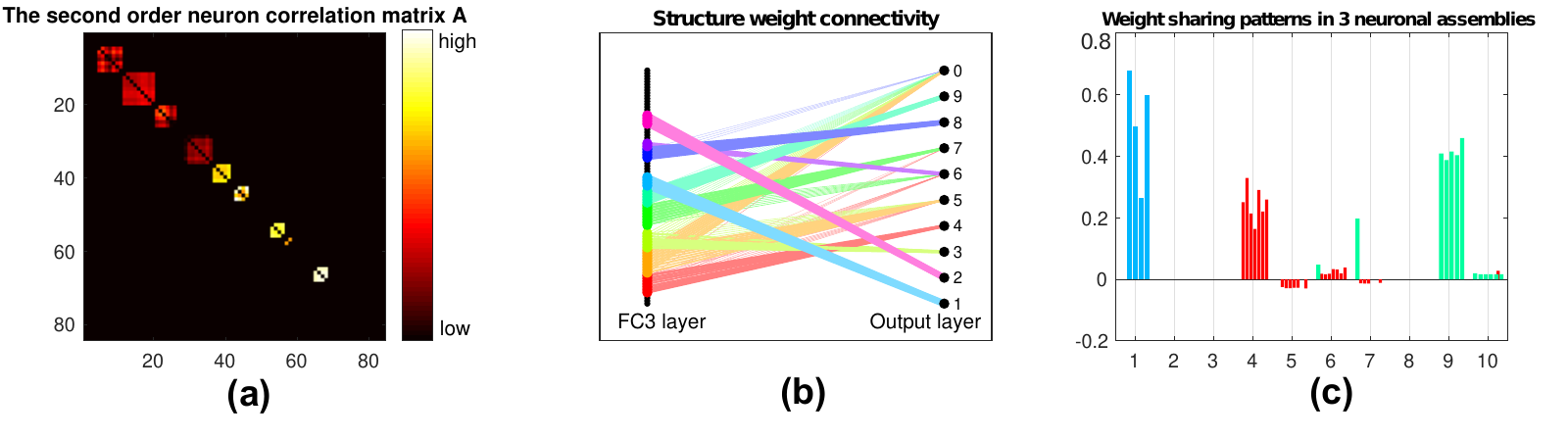}
\caption{Analysis and visualization of the last dense layer of LeNet-5+BEAN-2 model on the MNIST 10-shot learning from scratch task.
\textbf{(a)} Heat-map of the learned second-order neuron correlation matrix: neuron indices are re-ordered for best visualization of neuronal assembly patterns; BEAN is able to enforce plausible assembly patterns that act as functional clusters for the categorical learning task.
\textbf{(b)} Visualization of the parsimonious connectivity learned in the dense layer: both neuron-level and weight-level sparsity are simultaneously promoted in the network after applying BEAN regularization. The neurons are grouped and colored by neuronal assemblies.
\textbf{(c)} Visualization of the scales of neurons' outgoing weights: the weights of the neurons are colored to be consistent with the neuron group in (b).}
\label{fig:few_shot_study}
\end{figure}

Figure \ref{fig:few_shot_study} shows the learned parameters of the last dense layer of LeNet-5+BEAN2 on the MNIST 10-shot learning task. As shown in Figure \ref{fig:few_shot_study} (b), instead of using all possible weights in the dense layer, BEAN caused the model to parsimoniously leverage the weights and even the neurons, yielding a bio-plausible sparse and structured connectivity pattern. This is because the learned neuron correlation helped the model disentangle the co-connections between neurons from different assemblies, as shown in Figure \ref{fig:few_shot_study} (a). 
Additionally, BEAN enhanced parameter sharing among neurons within each assembly, as demonstrated in Figure \ref{fig:few_shot_study} (c). For instance, neurons in the red-colored assembly all had high positive weights toward class 4, meaning that this group of neurons was helping the model identify Digit 4. Similarly, neurons in the green-colored assembly were trying to distinguish between Digits 9 and 7. Such automatic weak parameter sharing not only helped prevent the model from over-fitting but also enabled an intuitive interpretation of the behavior of the system as a whole from a higher modularity level.

\begin{table}
  \caption{Statistic of data values test set error rate - validation set error rate on 10-shot learning on the MNIST dataset from 20 random runs. Other n-shot learning settings follow the same trend.}
  \label{tab:stat}
  \centering
  \begin{tabular}{c|cccccc}
    \toprule
    Model / Metric           &Max    &75\%-rank  &50\%-rank  &Mean   &25\%-rank  &Min\\
    \hline
    Vanilla                 &0.06\%   &0.01\%      &-0.06\%      &-0.30\% &-0.74\%     &-0.81\%\\
    \hline
    BEAN-1($\alpha=1$)      &-0.04\%   &-0.20\%      &-0.62\%       &-0.58\%  &-0.93\%        &-1.13\%\\
    BEAN-2($\alpha=100$)    &0.10\%     &-0.02\%        &-0.42\%      &-0.48\% &-0.97\%     &-1.35\%\\
    \bottomrule
  \end{tabular}
\end{table}

\subsubsection{Parameter sensitivity study}
There are two hyperparameters in the proposed BEAN regularization: 1) $\alpha$, which balances between the regularization loss and DNN training loss, and 2) $\gamma$, which controls the curvature of the hyperbolic tangent function as shown in Equation \ref{eq:W_hat}. As already mentioned in the first paragraph of Section 3, $\gamma$ was set to 1 for all experiments. Thus, the only parameter we need to study is $\alpha$.

\begin{figure}
\centering
\includegraphics[width=\linewidth]{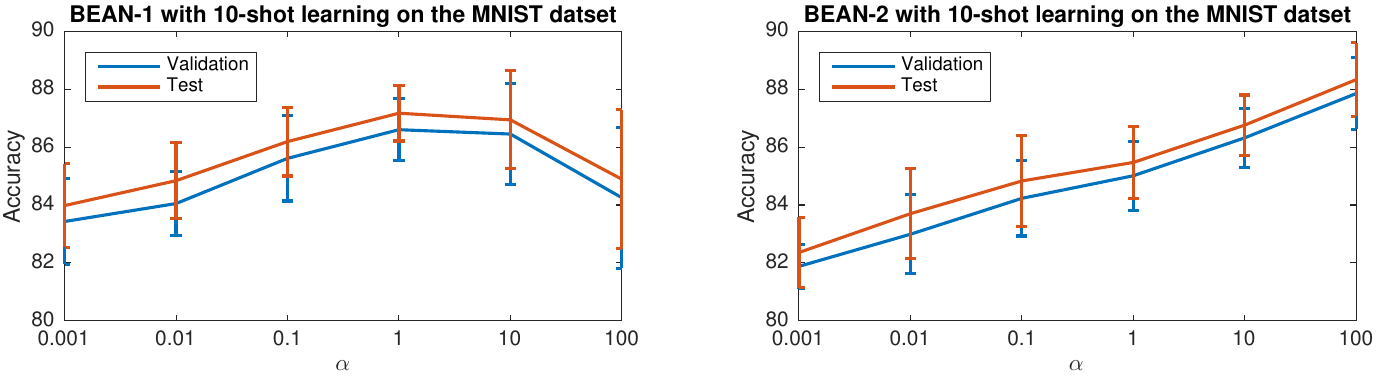}
\caption{Parameter sensitivity study of BEAN regularization on 10-shot learning on the MNIST dataset. Each data point is centered by the mean value and the error bar measures the standard deviation over 20 runs.}
\label{fig:sensitivity}
\end{figure} 

Figure \ref{fig:sensitivity} shows the accuracy of the model versus $\alpha$ on the few-shot learning setting on the MNIST dataset. Only the results for the 10-shot learning task are shown due to space limitations. By varying $\alpha$ across the range from 0.001 to 100, the best performance is obtained when $\alpha=1$ for BEAN-1 and $\alpha=100$ for BEAN-2. Specifically, for BEAN-1, We can see a clear trend where the model performance drops when $\alpha$ is too small or too big. Furthermore, the results show that the performance of the validation set is well aligned with the model performance on the test set, this demonstrates the superior generalizability of the model when applying BEAN regularization. 
Notably, although in Figure \ref{fig:sensitivity} we accessed the model performance on multiple settings of $\alpha$, we did not use any of the results on the test set to choose any parameters of the model, i.e. no post-selection was performed. We believe post-selection should be completely avoided and it can cause the test set to lose its power to test the model's generalizability to future unseen data.

\section{Conclusion}
In this paper, we propose a novel Biologically Enhanced Artificial Neuronal assembly (BEAN) regularization to model neuronal correlations and dependencies inspired by cell assembly theory from neuroscience.
We show that BEAN can promote jointly sparse and efficient encoding of rich semantic correlation among neurons in DNNs similar to connection patterns in BNNs.
Experimental results show that BEAN enables the formations of interpretable neuronal functional clusters and consequently promotes a sparse, memory/computation-efficient network without loss of model performance.
Moreover, our few-shot learning experiments demonstrated that BEAN could also enhance the generalizability of the model when training samples are extremely limited. 
Our regularization method has demonstrated its capability in enhancing the modularity of the representations of neurons for image semantic meanings such as digits, animals, and objects on image datasets. 
While the generality of the approach introduced here is at this time evaluated on MNIST and CIFAR datasets, future studies might consider additional experiments on other datasets such as texts or graphs to demonstrate the broader effectiveness of the proposed method. 
Another direction to further enhance the model might be to include separate excitatory and inhibitory nodes, as in BNNs, which would allow implementation of specific microcircuit computational motifs [\cite{ascoli2005incorporating}]. 
Furthermore, since there are other choices for defining the affinity matrix between neurons in a certain layer based on their downstream weights, answering the question about “what is the best way to compute affinity matrix” can be an interesting direction to be more comprehensively studied in future works. 

\section{ACKNOWLEDGMENTS}
This work is supported by the National Institutes of Health grant (NS39600), the National Science Foundation grant: \#1755850, \#1841520, \#1907805, Jeffress Trust Award, NVIDIA GPU Grant, and the Design Knowledge Company (subcontract number: 10827.002.120.04).
This manuscript has been released as a pre-print at arXiv: 1909.13698 [\cite{gao2019bean}].

\bibliographystyle{frontiersinSCNS_ENG_HUMS}
\bibliography{test}

\end{document}